\documentclass[letterpaper, 10 pt, conference]{ieeeconf}
\IEEEoverridecommandlockouts                             
\usepackage[x11names]{xcolor}

\usepackage{caption}
\usepackage{subcaption}
\usepackage{graphicx}
\usepackage{amsmath}
\usepackage{amssymb}
\usepackage{booktabs}
\usepackage{subfiles}
\usepackage{tabularx}
\usepackage{pifont}
\usepackage{color, colortbl}
\usepackage{algpseudocode,algorithm,algorithmicx}
\usepackage[11pt]{moresize}
\usepackage{tikz}
\usepackage{comment}
\usepackage{flushend}
\usepackage{verbatim}
\usepackage{multirow}
\usepackage{rotating}
\usepackage{hyperref}
\usepackage{microtype}
\usepackage{ifthen}
\usepackage[normalem]{ulem}
\usepackage{epstopdf}
\usepackage{tikz}

\usepackage{arydshln} 
\usepackage{array} 
\usepackage{makecell} 
\usepackage{xspace}
\usepackage{forloop}
\usepackage{lipsum}
\usepackage[noadjust]{cite}

\algnewcommand\algorithmicforeach{\textbf{for each}}
\algdef{S}[FOR]{ForEach}[1]{\algorithmicforeach\ #1\ \algorithmicdo}

\algnewcommand\algorithmicnot{\textbf{not}}
\algdef{SE}[IF]{IfNot}{EndIf}[1]{\algorithmicif\ \algorithmicnot\ #1\ \algorithmicthen}{\algorithmicend\ \algorithmicif}%

\definecolor{myGreen}{HTML}{E7FEE9}
\definecolor{myBlue}{HTML}{DFE5EF}
\definecolor{myOrange}{HTML}{FEF4D2}

\definecolor{vslamlab}{HTML}{800080} 
\definecolor{vslamlab_light}{HTML}{FBF5FB} 
\definecolor{orbslam}{HTML}{0000FF}
\definecolor{dpvo}{HTML}{FF0000}
\definecolor{droid}{HTML}{008000}
\definecolor{mast3r}{HTML}{FFA500}

\newcommand{\textpartone}{{\color{red}En un lugar de la Mancha, de cuyo nombre no quiero acordarme, no ha mucho tiempo que vivía un hidalgo de los de lanza en astillero, adarga antigua, rocín flaco y galgo corredor ...}}
\newcommand{\textparttwo}{{\color{red} Una olla de algo más vaca que carnero, salpicón las más noches, duelos y quebrantos los sábados, lantejas los viernes, algún palomino de añadidura los domingos, consumían las tres partes de su hacienda ...}}
\newcommand{\textpartthree}{{\color{red} El resto della concluían sayo de velarte, calzas de velludo para las fiestas, con sus pantuflos de lo mesmo, y los días de entresemana se honraba con su vellorí de lo más fino ...}}

\newcommand{\textpartfour}{{\color{red} Tenía en su casa una ama que pasaba de los cuarenta y una sobrina que no llegaba a los veinte, y un mozo de campo y plaza que así ensillaba el rocín como tomaba la podadera ...}}

\newcounter{mycounter}
\newcommand{\enUnLugar}{
\protect\stepcounter{mycounter}
\ifcase \themycounter%
    \or \textpartone%
    \or \textparttwo%
    \or \textpartthree%
    \or \textpartfour \setcounter{mycounter}{0}
    \else Text ended%
\fi
}

\newcolumntype{a}{:c|}
\newcolumntype{d}{:c}
\definecolor{tabfirst}{rgb}{0.4, 0.65, 0.3} 
\definecolor{tabsecond}{rgb}{0.7, 0.8, 0.65} 

\definecolor{nominal}{RGB}{27, 107, 162}
\definecolor{groundtruth}{RGB}{54, 161, 101}
\definecolor{repr_error}{RGB}{229, 126, 0}

\definecolor{proxy}{RGB}{187,136,187}

\newcommand{\numDatasets}[0]{12\xspace}  
\newcommand{\numPipelines}[0]{4\xspace}  
\newcommand{\numSequences}[0]{20\xspace}

\definecolor{lightblue}{rgb}{0.68, 0.85, 0.9} 

\newcommand{\myline}[1]{\textcolor{#1}{\raisebox{0.5ex}{\rule{0.4cm}{3.5pt}}}}
\newcolumntype{C}[1]{>{\centering\arraybackslash}m{#1}}

\newcommand{\dataset}[0]{dataset\xspace}

\makeatletter
\DeclareRobustCommand\onedot{\futurelet\@let@token\@onedot}
\def\@onedot{\ifx\@let@token.\else.\null\fi\xspace}

\def\eg{\emph{e.g}\onedot} 
\def\ie{\emph{i.e}\onedot}

\def\etal{\emph{et al}\onedot}
\makeatother

\usepackage{booktabs}
\usepackage{tikz}
\usepackage{graphicx}
\usepackage{listings}  
\usepackage{adjustbox}

\newcommand{\bluehref}[2]{\url{#1}}

\title{\LARGE \bf VSLAM-LAB: A Comprehensive Framework \\for Visual SLAM Methods and Datasets}

\author{Alejandro Fontan$^{\dag}$ \qquad Tobias Fischer$^{\dag}$ \qquad Javier Civera$^{\ddag}$ \qquad Michael Milford$^{\dag}$ \\ \bluehref{https://github.com/alejandrofontan/VSLAM-LAB}{https://github.com/alejandrofontan/VSLAM-LAB}  %
\thanks{$^{\dag}$AF, TF and MM are with the QUT Centre for Robotics, School of Electrical Engineering and Robotics, Queensland University of Technology, Brisbane, QLD 4000, Australia. $^{\ddag}$JC is with the I3A, Universidad de Zaragoza, Spain. This research was partially supported by funding from ARC Laureate Fellowship FL210100156 to MM and ARC DECRA Fellowship DE240100149 to TF. The authors acknowledge continued support from the Queensland University of Technology (QUT) through the Centre for Robotics. Corresponding author email:
        {\tt\small alejandro.fontan@qut.edu.au}}%
}

\begin{document}
\maketitle
\thispagestyle{empty}
\pagestyle{empty}
\begin{abstract}

Visual Simultaneous Localization and Mapping (VSLAM) research faces significant challenges due to fragmented toolchains, complex system configurations, and inconsistent evaluation methodologies. To address these issues, we present VSLAM-LAB, a unified framework designed to streamline the development, evaluation, and deployment of VSLAM systems. VSLAM-LAB simplifies the entire workflow by enabling seamless compilation and configuration of VSLAM algorithms, automated dataset downloading and preprocessing, and standardized experiment design, execution, and evaluation—all accessible through a single command-line interface. The framework supports a wide range of VSLAM systems and datasets, offering broad compatibility and extendability while promoting reproducibility through consistent evaluation metrics and analysis tools. By reducing implementation complexity and minimizing configuration overhead, VSLAM-LAB empowers researchers to focus on advancing VSLAM methodologies and accelerates progress toward scalable, real-world solutions. We demonstrate the ease with which user-relevant benchmarks can be created: here, we introduce difficulty-level-based categories, but one could envision environment-specific or condition-specific categories.
\end{abstract}

\section{INTRODUCTION}

Simultaneous Localisation and Mapping (SLAM) is a widely investigated problem within the computer vision and robotic communities \cite{cadena2016past}. It aims at estimating the trajectory of a mobile robot while concurrently constructing a representation of the environment. Visual SLAM (VSLAM) refers to the modality that utilizes RGB video as the main sensory input.

Despite significant research over the years, benchmarking VSLAM implementations remains highly challenging. On the one hand, VSLAM performance is highly dependent on the characteristics of the scene (\eg, its texture and depth distribution) and the camera motion (\eg, translation and linear and angular velocity profiles). This should motivate benchmarks in the widest variety of datasets, and including as many baselines as possible. However, the lack of standardization in datasets and implementations is a key barrier in doing this. Benchmarking is nowadays a time-consuming process for VSLAM researchers and practitioners and, as a result, experimental evaluations are typically limited to a small subset of sequences and baselines.

\begin{figure}[t]
\centering
\resizebox{\linewidth}{!}{ %
\begin{tabular}{cc}
\huge Easy  & \huge Medium  
\\
&
\\
\includegraphics[width=1\linewidth]{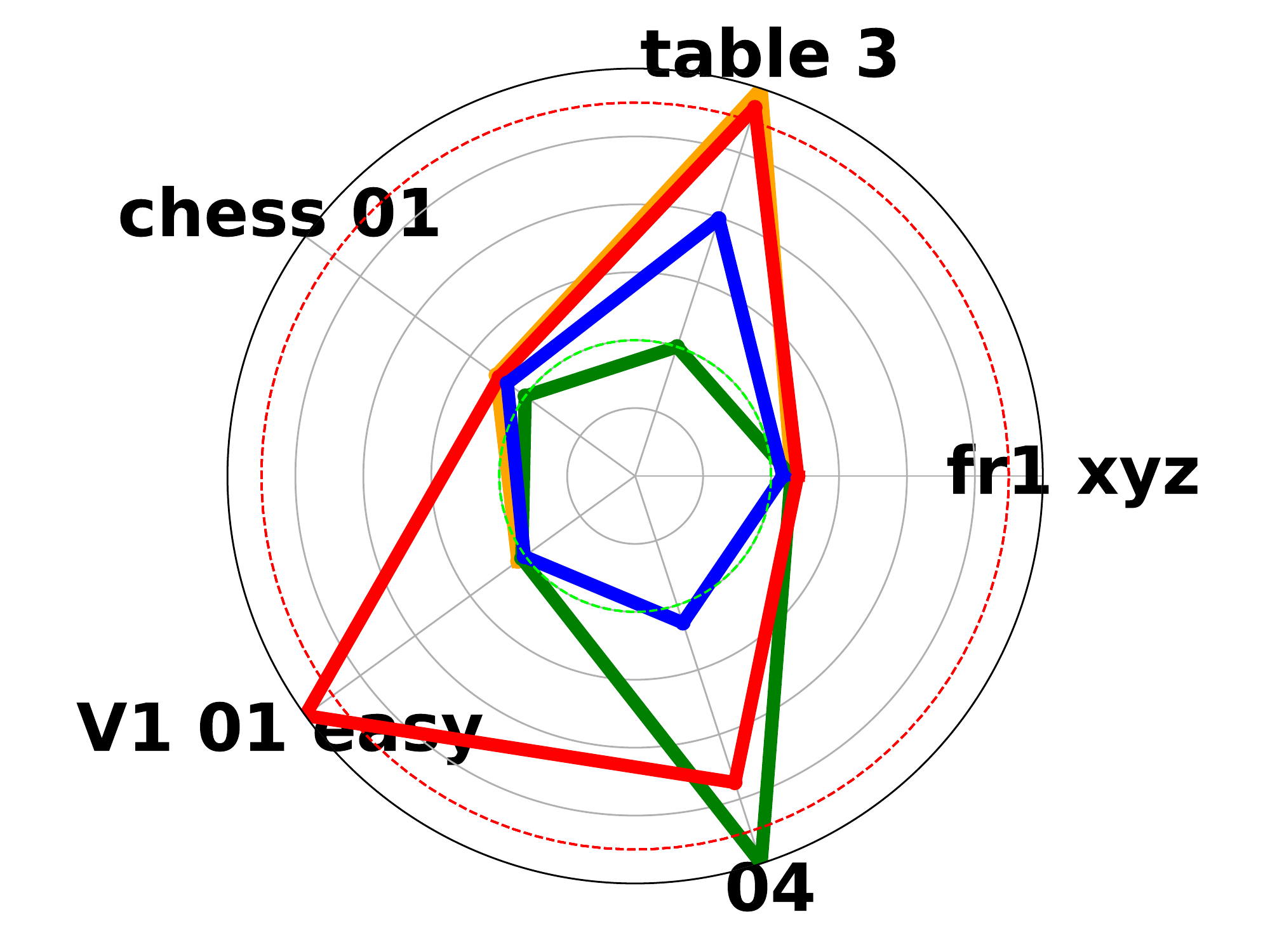}
&
\includegraphics[width=1\linewidth]{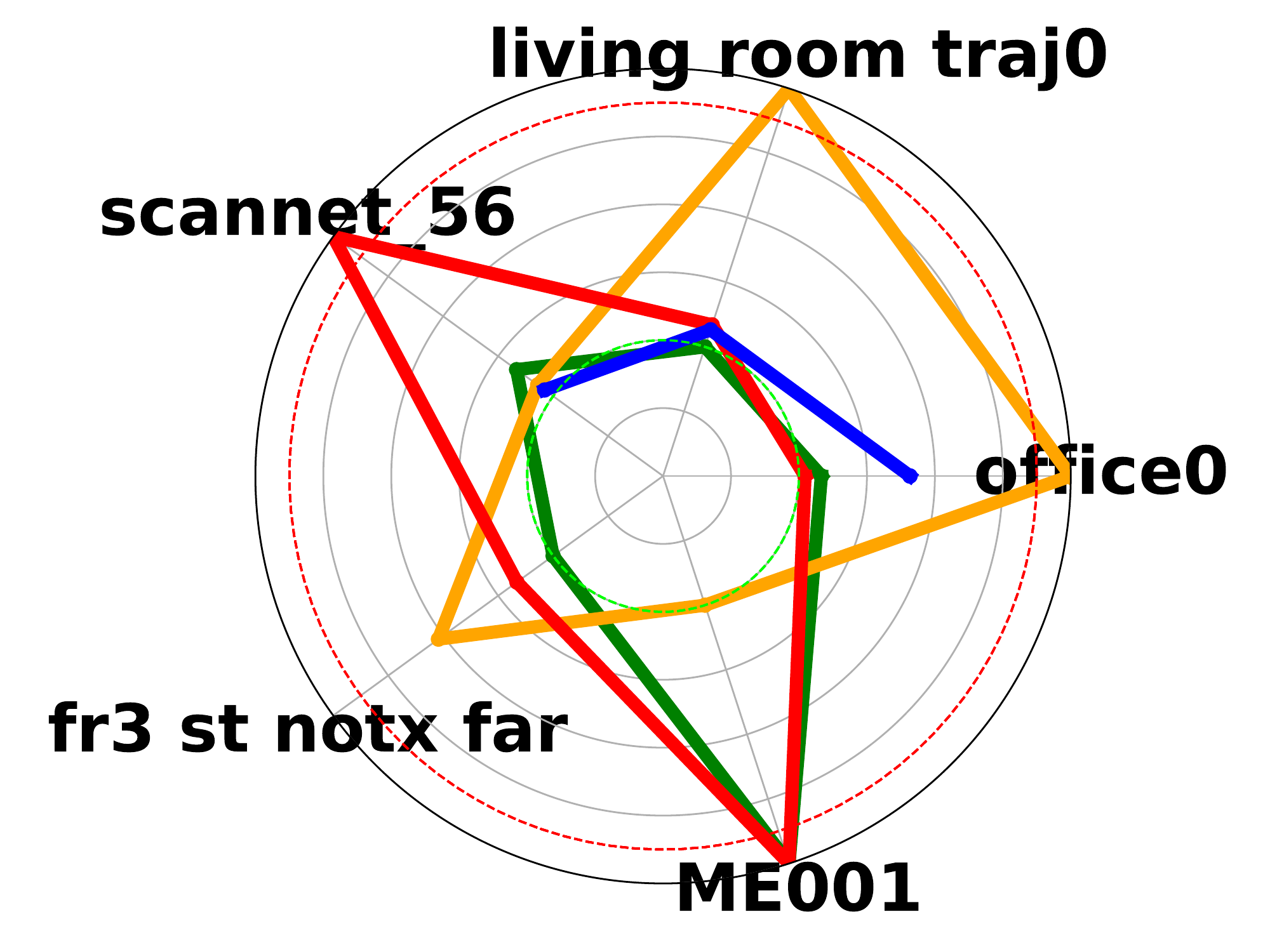} 
\\
&
\\
\huge Difficult  & \huge Extreme  
\\
&
\\
\includegraphics[width=1\linewidth]{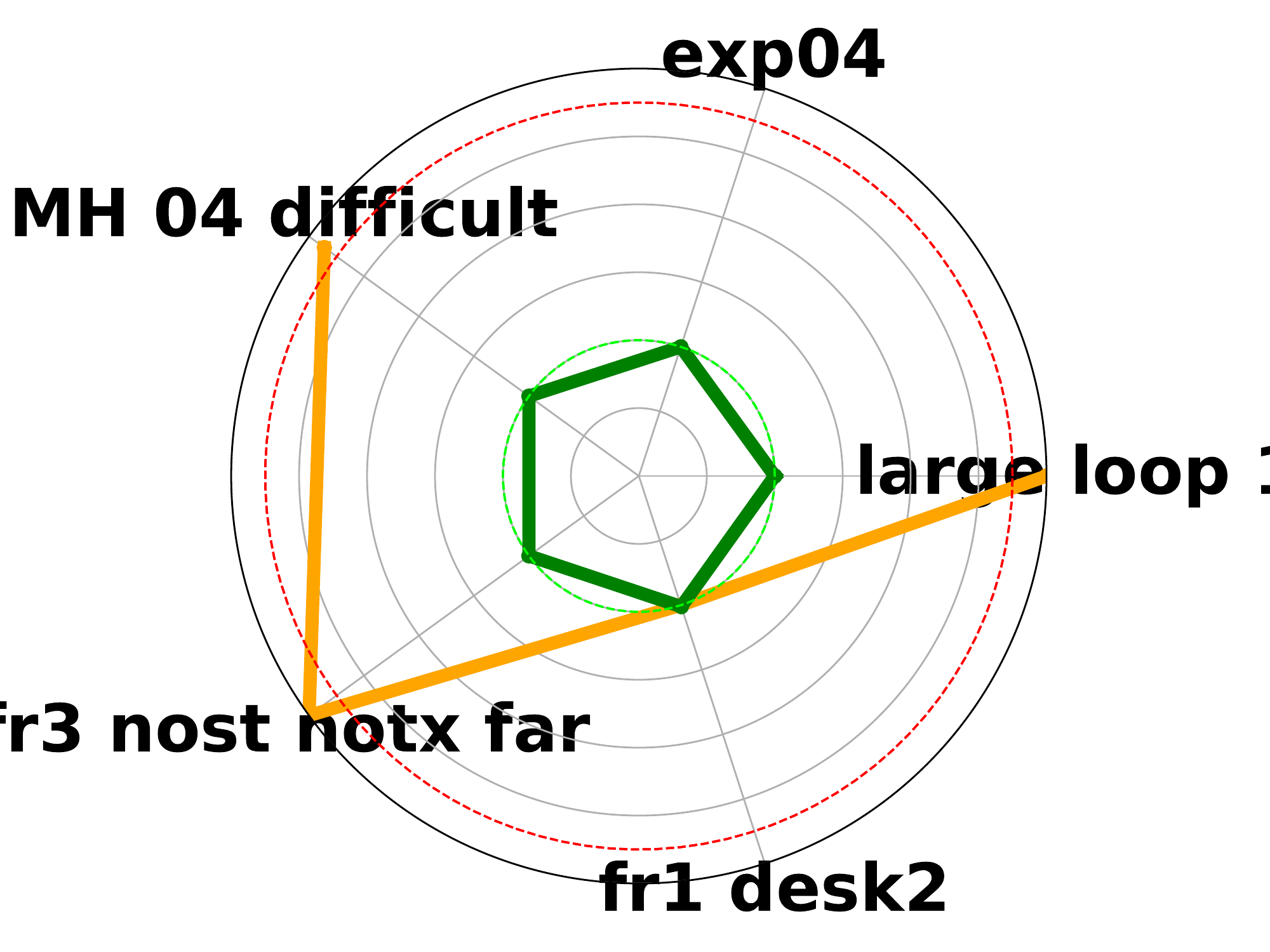} 
&
\includegraphics[width=1\linewidth]{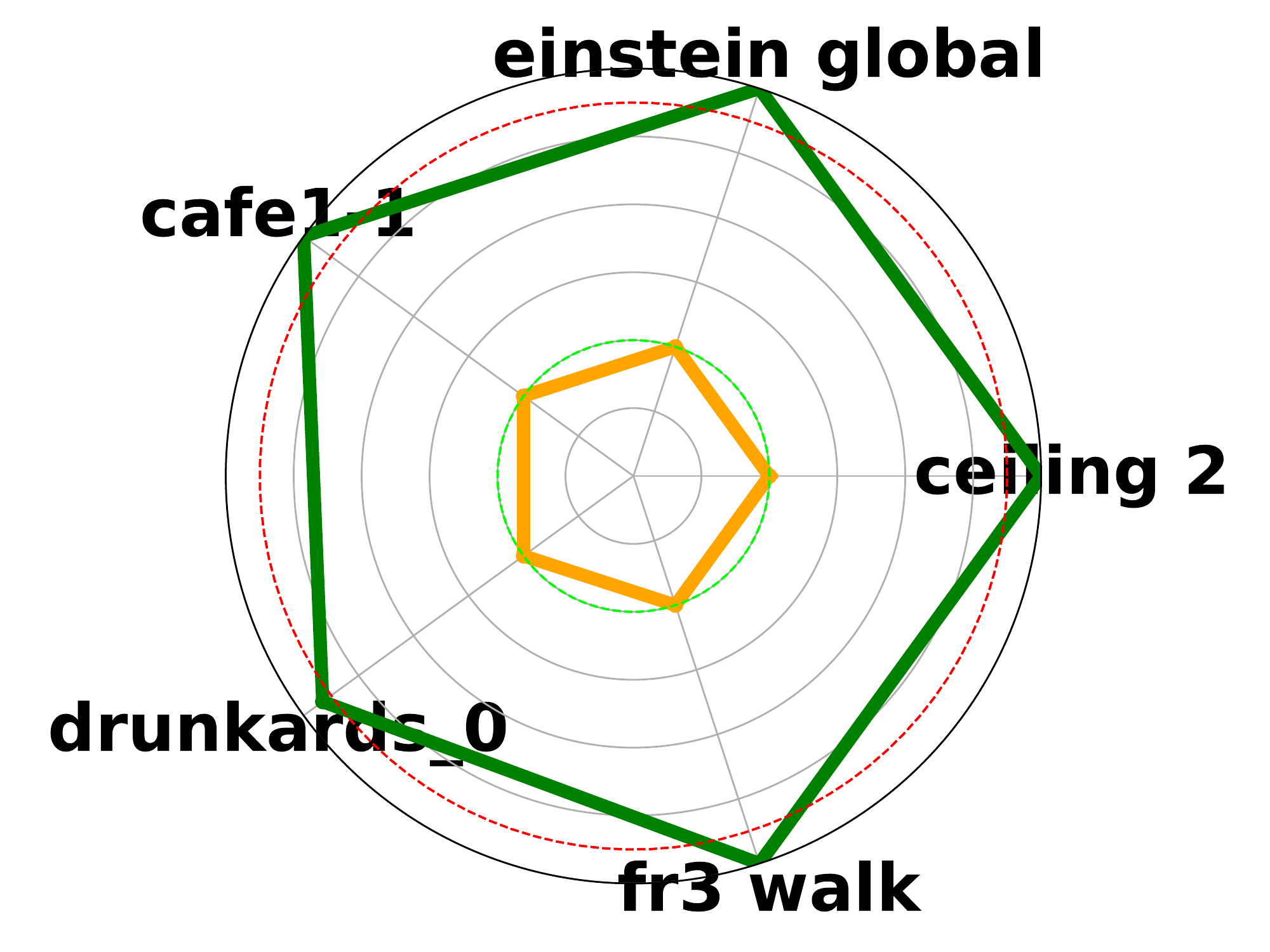} 
\end{tabular}
}
\resizebox{1\linewidth}{!}{
\begin{tabular}{ccccc}
\myline{mast3r} MASt3R-SLAM & \myline{droid} DROID-SLAM & \myline{dpvo} DPVO & \myline{orbslam} ORBSLAM2
\end{tabular}
}
\caption{Comparison of \numPipelines{} state-of-the-art Visual SLAM methods across \numSequences{} sequences from \numDatasets{} diverse datasets. Our evaluation highlights the strengths and weaknesses of each system, providing insights to guide future VSLAM research. Notably, the VSLAM-LAB framework ensures these experiments can be reproduced seamlessly with negligible time overhead and modified with minimal implementation effort. \textbf{Smaller area is better}.
}

\vspace*{-0.3cm}

\label{fig:radar_all}
\end{figure}
Fragmentation is indeed a major challenge in VSLAM research, with SLAM pipelines relying on disparate tools. Datasets differ in structure, camera calibration models, and ground truth formats, complicating direct comparisons. Inconsistencies in datasets, evaluation metrics, and benchmarking protocols further hinder reproducibility. Determining how trajectory ground truth is obtained—whether from independent motion-tracking systems like GPS or image-based reconstructions via structure-from-motion—remains time-consuming, as does automating parameter tuning~\cite{fontanRSS2024}. Additionally, many VSLAM implementations are not easily shareable or reusable and require complex dependencies and custom configurations, slowing research progress, limiting reproducibility, hindering comparisons, and complicating deployment and integration.

An underlying issue behind these challenges is that researchers often struggle with software complexities, writing code that addresses immediate needs but is difficult to reuse or reproduce. This results in significant time spent re-implementing evaluation scripts and formatting datasets, diverting focus from advancing VSLAM research.

To address these challenges, we introduce \textbf{VSLAM-LAB}—a powerful tool that streamlines benchmarking by automating key processes, including dataset formatting, method execution, experiment configuration, and trajectory evaluation. It enables effortless comparison of state-of-the-art VSLAM systems on standard and alternative datasets with minimal implementation effort while significantly reducing the time needed to integrate new methods and datasets. To ensure reproducibility, VSLAM-LAB employs a configuration file that standardizes the entire pipeline—from C++ and Python package installation to data formatting, execution, and evaluation—allowing experiments to be consistently replicated.

Here, we demonstrate the capabilities of \textbf{VSLAM-LAB} by evaluating \numPipelines{} state-of-the-art VSLAM pipelines across \numSequences{} sequences from \numDatasets{} diverse datasets. Furthermore, we introduce the first benchmark that integrates challenges from all these datasets, encompassing a wide range of scenarios, including indoor and outdoor environments, varying difficulty levels, real and synthetic data, and scenes with dynamic objects. VSLAM-LAB enables researchers to effortlessly create custom benchmarks tailored to specific research questions. While we showcase difficulty-level categorization as one example (see Figure~\ref{fig:radar_all}), VSLAM-LAB's flexible architecture supports many other classification schemes—such as environment type, motion patterns, or lighting conditions—without requiring significant implementation effort.

\section{RELATED WORK}
In this section, we first briefly summarize the most popular SLAM datasets in Section~\ref{subsec:VSLAMdatasets}, followed by other works that aim toward standardised benchmarking in SLAM in \mbox{Section~\ref{subsec:VSLAMbenchmarking}}. We also review a range of localization frameworks that share our aims of standardizing dataset formatting, method execution and evaluation in Section~\ref{subsec:localization}.

\subsection{Visual SLAM Datasets}
\label{subsec:VSLAMdatasets}
The TUM RGB-D benchmark~\cite{sturm12iros} by Sturm \etal~provides a comprehensive dataset of 39 sequences captured with a Microsoft Kinect in office and industrial environments. The benchmark includes synchronized color and depth images at 640×480 resolution with ground truth camera poses from a motion capture system.

The ETH3D SLAM benchmark by Schöps \etal~\cite{schops2017multi,schops2019bad}~features 56 training and 35 test datasets recorded with a custom camera rig. Similarly to the TUM RGB-D benchmark, the ground truth annotations were obtained using a motion capture system, with some additional sequences whose ground truth was approximated via Structure-from-Motion. The ETH3D SLAM benchmark investigates performance impacts due to rolling shutter and geometric distortions in cameras. Interestingly, their benchmark contains sequences without public ground truth to avoid overfit to specific datasets/sequences.

The KITTI vision benchmark suite~\cite{geiger2012we} represents a pioneering effort in autonomous driving datasets, featuring stereo, optical flow, visual odometry/SLAM, and 3D object detection tasks. Captured using a sensor-rich autonomous driving platform, the benchmark highlighted the increased difficulty of real-world scenarios compared to controlled lab conditions.

Engel \etal \cite{engel2016photometrically, engel2017direct} presented the TUM monoVO dataset, designed to offer photometrically calibrated image sequences recorded in a variety of indoor and outdoor settings. The dataset prioritizes camera motion analysis, incorporating a significant loop closure at the end of each sequence to assess drift accumulation without relying on complete ground truth trajectories. Visual odometry (VO) performance is evaluated using the alignment error, a metric that quantifies drift across the entire sequence.

\subsection{Visual SLAM Benchmarking Suites}
\label{subsec:VSLAMbenchmarking}

The pySLAM framework~\cite{freda2025pyslam} provides a Python implementation of visual SLAM supporting multiple camera configurations. It offers various local features, loop closing methods, and a volumetric reconstruction pipeline. While pySLAM provides implementation tools for SLAM algorithms including a large range of local feature detectors and descriptors, VSLAM-LAB differs by focusing on the comprehensive evaluation ecosystem, addressing the fragmentation in evaluation methodologies that has hindered objective comparison of SLAM approaches. PySLAM is also limited to currently just four datasets.

The evo Python package~\cite{grupp2017evo} provides functionality for comparing and evaluation trajectory outputs from odometry and SLAM algorithms. It supports multiple trajectory formats including TUM, KITTI, and EuRoC MAV, as well as ROS and ROS2 bagfiles. While evo provides valuable tools for trajectory analysis, it primarily focuses on the evaluation stage.

SLAMBench~\cite{nardi2015introducing,Bodin2018,bujanca2019slambench,bujanca2021robust} may be the closest to our work. It is, however, not maintained currently, the most recent VSLAM baselines (\eg, DPVO, DROID-SLAM or MASt3R-SLAM) being missing. The SLAM Hive Suite~\cite{yang2023slam} shares the same target, but operates using Docker containers, hence enabling parallel benchmarking in the cloud. It does not, however, address standardization, and hence still may incur in programming overhead for adapting systems, datasets and benchmarks. In this sense, it is complementary to our work, and utilizing it downstream to parallelize different runs from our VSLAM-LAB would be interesting as future work.

While these benchmarks offer valuable insights, they primarily focus on isolated use cases, limiting their generalizability to broader SLAM applications. VSLAM-LAB overcomes this issue by offering a diverse collection of datasets across indoor, outdoor, synthetic, and challenging environments. It also provides a standardized evaluation process, enabling researchers to compare SLAM methods more systematically and under various real-world conditions.

\subsection{Localization Frameworks}
\label{subsec:localization}
Various localization benchmark efforts share our goal of standardizing evaluation methodologies, but target different technical challenges within the broader visual perception domain.

The deep visual geo-localization benchmark~\cite{berton2022deep} provides a framework for researchers to build, train, and test various geo-localization architectures with modular components. This framework emphasizes performance metrics such as recall@N alongside system requirements including execution time and memory consumption. While sharing our objective of standardizing evaluation protocols, this work focuses specifically on geo-localization rather than the full SLAM pipeline that VSLAM-LAB addresses.

Complementary to the deep visual geo-localization benchmark~\cite{berton2022deep}, the visual localization benchmark~\cite{toft2020long} provides several benchmark datasets for 6-DoF pose estimation. The datasets cover a range of appearance changes, including seasonal, viewpoint and illumination (dawn, day, sunset, night) conditions. Interestingly, the query poses are withheld to avoid overfitting of methods on specific datasets.

Several works provide comprehensive frameworks for evaluation visual place recognition (VPR), which is often used as a loop closure component for SLAM. VPR-Bench~\cite{zaffar2021vpr} offers 12 integrated datasets and benchmarks 10 VPR techniques, with particular emphasis on quantifying viewpoint and illumination invariance. The VPR methods evaluation codebase~\cite{Berton_2023_EigenPlaces} provides access to many state-of-the-art VPR techniques including the model definitions and weights, and enables researchers to compare them fairly. It complements the VPR Datasets Downloader repository by the same authors, which similarly to VSLAM-LAB provides a unified way to access datasets in a standardized way, in this case for VPR datasets~\cite{berton2022deep}.

Despite progress in SLAM research, localization benchmarking remains a largely manual process, increasing the risk of inconsistencies and reducing reproducibility. VSLAM-LAB offers a fully integrated solution that standardizes localization evaluation, ensuring that different methods can be fairly and efficiently compared while reducing the complexity of setting up and running experiments.

\begin{table}[t]
\footnotesize
\centering
\setlength{\tabcolsep}{5pt} 
\resizebox{1.0\columnwidth}{!}{
\begin{tabular}{ccccccccccccccccccccc}
\arrayrulecolor{vslamlab}
2016 & 2017  & 2021 & 2024  & & & \cellcolor{gray!15} & \cellcolor{gray!15}  & \cellcolor{gray!15} & 2025 \\
\midrule
\rotatebox{270}{\href{https://github.com/JakobEngel/dso}{DSO}~\cite{engel2017direct} } &
\rotatebox{270}{\href{https://github.com/raulmur/ORB_SLAM2}{ORB-SLAM2}~\cite{mur2017orb}} &
\rotatebox{270}{\href{https://github.com/princeton-vl/DROID-SLAM}{DROID-SLAM}~\cite{teed2021droid} } &
\rotatebox{270}{\href{https://github.com/alejandrofontan/AnyFeature-VSLAM}{AnyFeature-VSLAM}~\cite{fontanRSS2024} } &
\rotatebox{270}{\href{https://github.com/muskie82/MonoGS}{MonoGS}~\cite{matsuki2024gaussian}} &
\rotatebox{270}{\href{https://github.com/princeton-vl/DPVO}{DPVO}~\cite{lipson2025deep}} &
\rotatebox{270}{ \cellcolor{gray!15} \href{https://github.com/colmap/glomap}{GLOMAP}~\cite{pan2024global}} &
\rotatebox{270}{\cellcolor{gray!15} \href{https://github.com/naver/dust3r}{DUSt3R}~\cite{wang2024dust3r}} & 
\rotatebox{270}{\cellcolor{gray!15} \href{https://github.com/HengyiWang/spann3r}{Spann3R}~\cite{wang2024dust3r}} &
\rotatebox{270}{\href{https://github.com/rmurai0610/MASt3R-SLAM}{MASt3R-SLAM}~\cite{murai2024mast3r}} &
\\
\bottomrule
\end{tabular}
}
\caption{\textbf{VSLAM-LAB Methods}: Visual SLAM \& Multi-view reconstruction methods (gray columns) for monocular cameras contained in VSLAM-LAB.}
\label{tab:pipelines}
\end{table}

\begin{table}[t]
\footnotesize
\centering
\setlength{\tabcolsep}{5pt} 
\resizebox{1.0\columnwidth}{!}{
\begin{tabular}{ccccccccccccccccccccc}
\arrayrulecolor{vslamlab}
2012 &  & 2013 & \cellcolor{gray!15} 2014 & 2016 & 2017 & 2018 & 2019 & \cellcolor{gray!15} & 2020 & \cellcolor{gray!15}   & 2021 & 2022& 2023 &   & \cellcolor{gray!15} \\
\midrule
\rotatebox{270}{\href{https://cvg.cit.tum.de/data/datasets/rgbd-dataset}{TUM-RGBD}~\cite{sturm12iros}} &
\rotatebox{270}{\href{https://www.cvlibs.net/datasets/kitti/eval_odometry.php}{KITTI}~\cite{geiger2013vision}} &
\rotatebox{270}{\href{https://www.microsoft.com/en-us/research/project/rgb-d-dataset-7-scenes/}{7-Scenes}~\cite{glocker2013real}} &
\rotatebox{270}{\href{https://www.doc.ic.ac.uk/~ahanda/VaFRIC/iclnuim.html}{ICL-NUIM \cellcolor{gray!15}}~\cite{handa2014benchmark}} &
\rotatebox{270}{\href{https://projects.asl.ethz.ch/datasets/doku.php?id=kmavvisualinertialdatasets}{EuRoC}~\cite{yeshwanth2023scannet++}} &
\rotatebox{270}{\href{https://cirs.udg.edu/caves-dataset/}{Caves}~\cite{mallios2017underwater}} &
\rotatebox{270}{\href{https://datasets.arches-projekt.de/morocco2018/}{MADMAX}~\cite{meyer2021madmax}} &
\rotatebox{270}{\href{https://www.eth3d.net/slam_datasets}{ETH3D}~\cite{schops2019bad}} &
\rotatebox{270}{\href{https://github.com/facebookresearch/Replica-Dataset}{\cellcolor{gray!15} Replica}~\cite{replica19arxiv}} & 
\rotatebox{270}{\href{https://lifelong-robotic-vision.github.io/dataset/scene.html}{OpenLORIS}~\cite{shi2020we}} &
\rotatebox{270}{\href{https://theairlab.org/tartanair-dataset/}{\cellcolor{gray!15} TartanAir}~\cite{wang2020tartanair}} &
\rotatebox{270}{\href{https://davidrecasens.github.io/EndoDepthAndMotion/}{Hamlyn}~\cite{recasens2021endo, mountney2010three}} &
\rotatebox{270}{\href{https://lamar.ethz.ch/}{LaMAR}~\cite{sarlin2022lamar}} &
\rotatebox{270}{\href{https://kaldir.vc.in.tum.de/scannetpp/}{Scannet++}~\cite{yeshwanth2023scannet++}}  &
\rotatebox{270}{\href{https://hilti-challenge.com/dataset-2022.html}{HILTI 2022}~\cite{zhang2022hilti}} &
\rotatebox{270}{\href{https://davidrecasens.github.io/TheDrunkard'sOdometry/}{\cellcolor{gray!15} Drunkard's}~\cite{recasens2023drunkard}} 
 \\
\bottomrule
\end{tabular}
}
\caption{\textbf{VSLAM-LAB Datasets:} The white columns correspond to datasets contained in our VSLAM-LAB capturing real-world data, while the gray columns contain synthetic datasets.}
\label{tab:datasets}
\end{table}

\section{VSLAM-LAB}
VSLAM-LAB provides a flexible framework for conducting customizable experiments with minimal configuration. It offers an automated pipeline for method compilation and installation (\ref{subsec:dependencymanagement}), supports seamless integration of various methods (Section~\ref{subsec:methods} with many datasets (\ref{subsec:datasets}), \dataset acquisition and preprocessing, experiment execution (Section~\ref{subset:experimentcustomization}), and comprehensive evaluation (Section~\ref{subsec:evaluation}).

\subsection{Dependency management}
\label{subsec:dependencymanagement}
A key aspect of VSLAM-LAB's implementation is the use of \emph{Pixi} (\url{https://pixi.sh}) for dependency management. Pixi is a multi-platform and multi-language package management tool that extends the popular Conda ecosystem. In addition to seamlessly installing compilers and low-level libraries like CUDA, it provides binary packages for many popular tools like PyTorch, OpenCV, Open3D, Scikit-Learn and many more.

VSLAM-LAB specifies separate environments for each of the VSLAM methods listed in the next section, and the environments' dependencies are listed in the \textit{pixi.toml} manifest file. Importantly for reproducibility, pixi automatically creates lock files for all dependencies, which ensures the creation of consistent environments, guaranteeing reproducibility across different setups by enforcing identical package versions and dependencies. Pixi makes running an experiment on a new machine as easy as:
\begin{lstlisting}
# install pixi
curl -fsSL https://pixi.sh/install.sh | bash
# clone repo
git clone https://github.com/alejandrofontan/VSLAM-LAB.git
# run experiment, e.g. our demo
cd VSLAM-LAB && pixi run demo
\end{lstlisting}

\subsection{Visual SLAM Methods}
\label{subsec:methods}
Table~\ref{tab:pipelines} (top) lists the Visual SLAM methods currently integrated into VSLAM-LAB. We include four dense methods, \ie MASt3R-SLAM~\cite{murai2024mast3r}, DPVO~\cite{lipson2025deep}, MonoGS~\cite{matsuki2024gaussian}, and DROID-SLAM~\cite{teed2021droid}, and three sparse visual SLAM methods, \ie AnyFeature-VSLAM~\cite{fontanRSS2024}, ORB-SLAM2~\cite{mur2017orb}, and DSO~\cite{engel2017direct}. Among these, MASt3R-SLAM is the only approach capable of handling sequences captured with uncalibrated cameras.

A common strategy for evaluating Visual SLAM methods in the absence of ground-truth data is to generate a pseudo-ground-truth using an offline structure-from-motion technique. VSLAM-LAB integrates GLOMAP~\cite{pan2024global}, enabling direct comparison with offline approaches. Furthermore, VSLAM-LAB includes multi-view stereo reconstruction methods DUSt3R~\cite{wang2024dust3r} and Spann3R~\cite{wang2024spann3r} (Table~\ref{tab:pipelines}, bottom).

\subsubsection*{Adding a New Method} To incorporate a new method, users must first specify the method's dependencies in the \textit{pixi.toml} manifest. Users also need to define an \emph{install} task, which typically runs \emph{pip} for Python packages or builds a C++ package with \emph{cmake}.

After the dependencies are defined and the method is installed, users then need to define a new class derived from \textit{BaselineVSLAMLab} and in the simplest case specify the method's name, folder, and default parameter, as shown in the code listing below. The newly created class then needs to be added to the list of available methods in \textit{baseline\_utilities.py}. An example class implementing DSO is as follows:
\begin{lstlisting}
class DSO_baseline(BaselineVSLAMLab):
    def __init__(self):
        baseline_name = "dso"
        baseline_folder = "dso"
        default_parameters = ["Preset: preset:0", "Mode: mode:1"]

        # Initialize the baseline
        super().__init__(baseline_name, baseline_folder, default_parameters)
\end{lstlisting}

VSLAM-LAB executes the script specified in the \textit{execute} task within the \textit{pixi.toml} manifest, which runs the method while processing the necessary inputs (see Section~\ref{subsec:datasets}); specifically:
\begin{lstlisting}
--sequence_path  # Path Sequence
--calib_yaml     # Path calibration.yaml
--rgb_csv        # Path rgb.csv
--exp_id         # e.g. 00000 
--settings_yaml  # Path method_set.yaml
--visualization  # True / False
\end{lstlisting}

The method must output a trajectory with a rigid transformation per frame in the required format (see Section~\ref{subsec:datasets}).

\subsection{Visual SLAM Datasets} 
\label{subsec:datasets}

Table~\ref{tab:datasets} lists all datasets available through VSLAM-LAB. Each dataset is automatically downloaded and converted into a standardized format to facilitate downstream processing. The directory structure is as follows, loosely following the TUM-RGBD Benchmark:

\begin{verbatim}
VSLAMLAB-BENCHMARK/DATASET/Sequence
|-- rgb/
|   |-- rgb_0000.jpg
|   |-- rgb_0001.jpg
|   |-- rgb_0002.jpg
|   |-- ...
|-- rgb.csv
|-- groundtruth.csv
|-- calibration.yaml
\end{verbatim}

\textbf{Trajectory Format:} All trajectory files in VSLAM-LAB adhere to the format defined in~\cite{sturm12iros}:  
\texttt{ts tx ty tz qx qy qz qw},  
where \texttt{ts} represents the timestamp, \texttt{tx, ty, tz} are the translation components, and \texttt{qx, qy, qz, qw} define the orientation as a quaternion.

\textbf{Image Undistortion:} As part of the dataset preprocessing, all RGB images are undistorted to conform to a pinhole camera model. This ensures a consistent data format across all methods, eliminating the need for additional undistortion steps during evaluation.

\subsubsection*{Adding a New Dataset} Similarly to adding a new method, to incorporate a new dataset, users must create a new class derived from \textit{DatasetVSLAMLab} and add it to \textit{dataset\_utilities}. The class functions should be overridden to automate data downloading and formatting according to the required structure. Specifically, the \textit{download\_sequence\_data()} method should download and uncompress the dataset files, the \textit{create\_rgb\_folder()} method should arrange the files according to the format specified above, the \textit{create\_calibration\_yaml()} should create the \textit{calibration.yaml} in OpenCV format, the \textit{create\_rgb\_txt()} method should create the \textit{rgb.csv} that lists all images in the correct order, and finally the \textit{create\_groundtruth\_txt()} method should create the \textit{groundtruth.csv} adhering to the trajectory format detailed above.

\subsection{Experiment Customization} 
\label{subset:experimentcustomization}
VSLAM-LAB allows for easy customization of experiments through a simple configuration process. Users can select specific methods, define the number of runs (considering non-deterministic outputs of most methods), and specify input parameters for each method to fine-tune its performance. The configuration file follows this structure:  

\begin{lstlisting}
# VSLAMLAB/configs/exp_config_easy.yaml

exp_config_easy_mast3rslam:
  Config: config_easy.yaml
  NumRuns: 10
  Parameters: {verbose: 0, max_rgb: 120}
  Method: mast3rslam

exp_config_easy_dpvo:
  Config: config_easy.yaml
  NumRuns: 10
  Parameters: {verbose: 0, max_rgb: 120}
  Method: dpvo
\end{lstlisting}

Additionally, users can customize the datasets used in an experiment by specifying sequences from different datasets within the configuration file:

\begin{lstlisting}
# VSLAMLAB/configs/config_easy.yaml

rgbdtum:
- freiburg3_structure_texture_far
eth:
- table_3
7scenes:
- chess_seq-01
euroc:
- V1_01_easy
\end{lstlisting}

\subsubsection*{Adding a New Configuration} New configurations can be easily introduced by creating a new \textit{config} file. This approach ensures continuous adaptability as methods evolve and datasets increase in complexity and diversity. Moreover, it enhances reproducibility by enabling researchers to replicate specific experiments simply by sharing the corresponding configuration file.

\subsection{Evaluation}
\label{subsec:evaluation}

\textbf{Absolute Trajectory Error (ATE)}~\cite{sturm12iros}. Since monocular visual SLAM produces unscaled trajectories, ATE serves as the standard evaluation metric. It measures global consistency by aligning estimated and ground-truth trajectories~\cite{horn1987closed,umeyama1991least} and computing translational differences.

\begin{figure*}[t]
    \centering
    \begin{subfigure}[b]{0.49\textwidth}
        \centering
        \resizebox{\linewidth}{!}{ %
            \begin{tabular}{C{0.15\linewidth}C{0.15\linewidth}C{0.15\linewidth}C{0.15\linewidth}C{0.15\linewidth}C{0.15\linewidth}}
            \arrayrulecolor{vslamlab}
            \footnotesize RGBDTUM& \footnotesize ETH3D & \footnotesize 7SCENES & \footnotesize EUROC  &  \footnotesize KITTI \\
            \footnotesize Fr.1 xyz & \footnotesize Table 3 & \footnotesize Chess 01 & \footnotesize V1 01 E.&  \footnotesize KITTI 04 \\
            \hline
            \\
            \multicolumn{5}{c}{\includegraphics[width=1\linewidth]{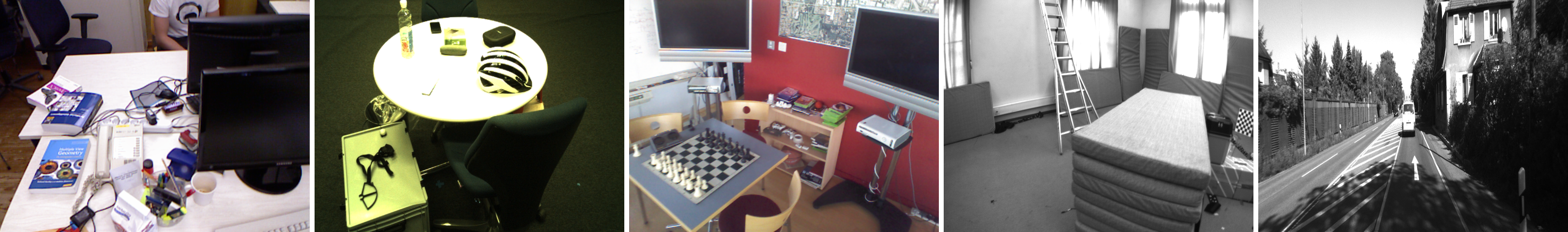}} \\
            \multicolumn{5}{c}{\includegraphics[width=1\linewidth]{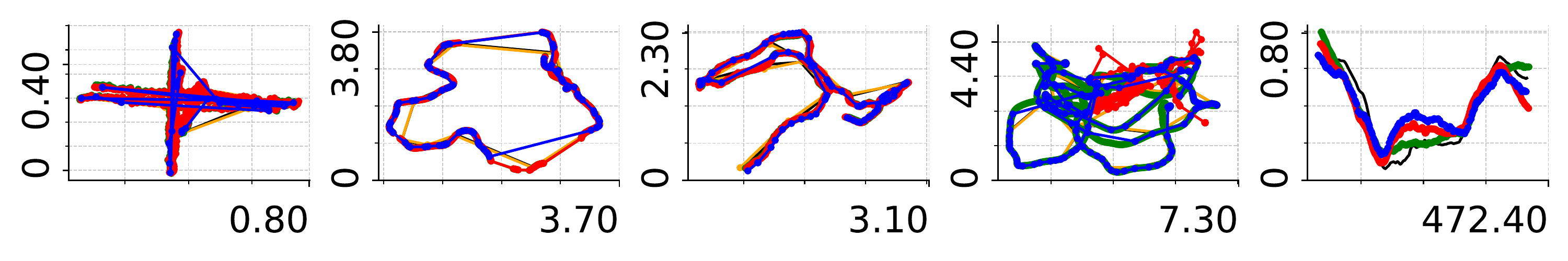}} \\
            \multicolumn{5}{c}{\includegraphics[width=1\linewidth]{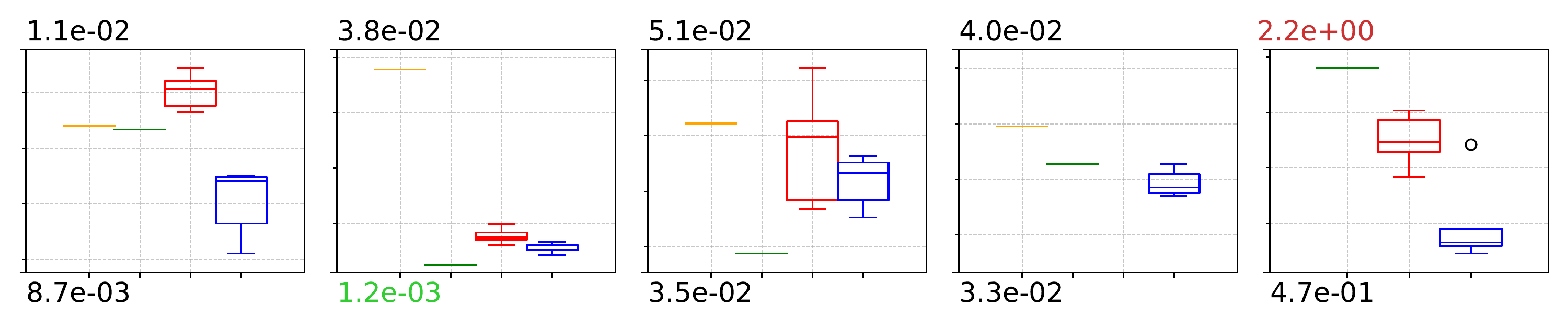}} \\
            \end{tabular}
        }
        \resizebox{1\linewidth}{!}{
            \begin{tabular}{cccc}
            \myline{mast3r} MASt3R-SLAM & \myline{droid} DROID-SLAM & \myline{dpvo} DPVO & \myline{orbslam} ORB-SLAM2
            \end{tabular}
        }
        \caption{\textbf{Easy 2025}}
        \label{fig:configEasy}
    \end{subfigure}
    \hfill
    \begin{subfigure}[b]{0.49\textwidth}
        \centering
        \resizebox{\linewidth}{!}{ %
        \begin{tabular}{C{0.15\linewidth}C{0.15\linewidth}C{0.15\linewidth}C{0.15\linewidth}C{0.15\linewidth}C{0.15\linewidth}}
        \arrayrulecolor{vslamlab}
        \footnotesize REPLICA& \footnotesize ICLNUIM & \footnotesize SCANNET++ & \footnotesize RGBDTUM  &  \footnotesize TARTANAIR \\
        \footnotesize Office 0 & \footnotesize l.r. traj0 & \footnotesize Scannet 56 & \footnotesize Fr.3 nsnt f. &  \footnotesize ME001 \\
        \hline
        \\
        \multicolumn{5}{c}{\includegraphics[width=1\linewidth]{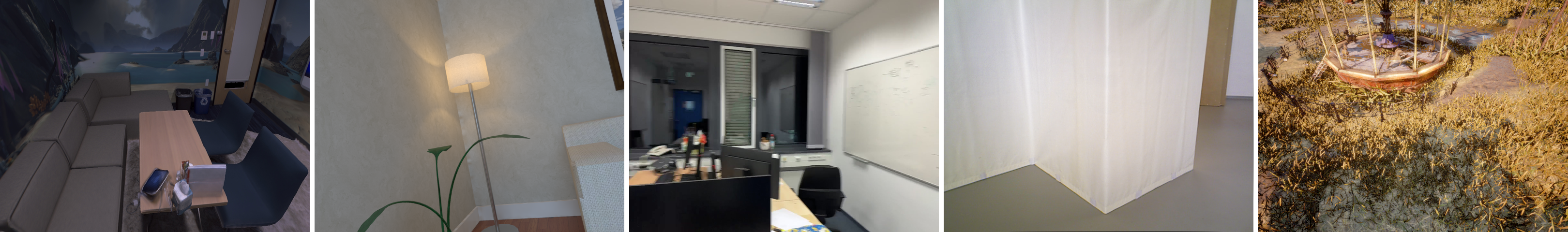}} \\
        \multicolumn{5}{c}{\includegraphics[width=1\linewidth]{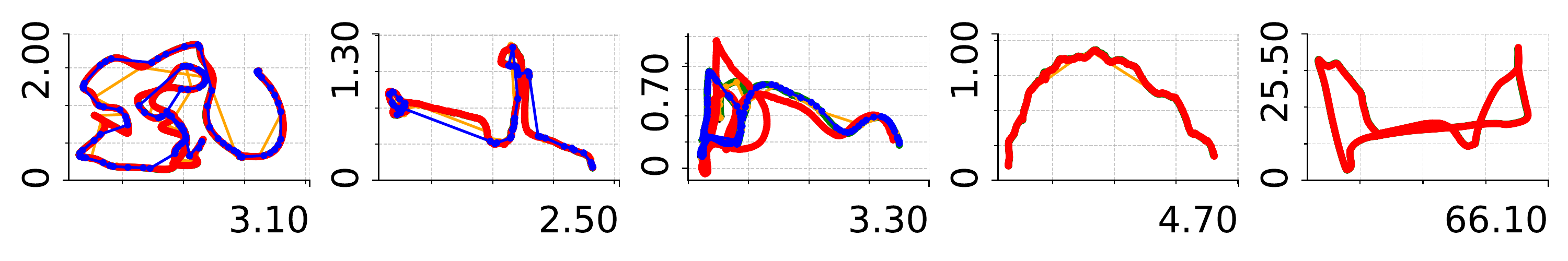}} \\
        \multicolumn{5}{c}{\includegraphics[width=1\linewidth]{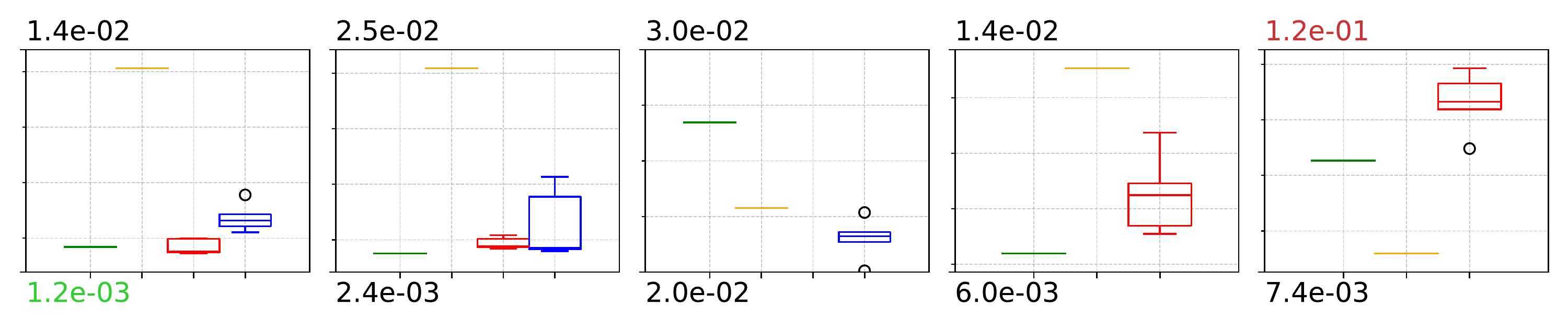}} \\
        \end{tabular}
        }
        \resizebox{1\linewidth}{!}{
            \begin{tabular}{cccc}
            \myline{mast3r} MASt3R-SLAM & \myline{droid} DROID-SLAM & \myline{dpvo} DPVO & \myline{orbslam} ORB-SLAM2
            \end{tabular}
        }
        \caption{\textbf{Medium 2025}}
        \label{fig:configMedium}
    \end{subfigure}

    \vspace{0.5cm}
    
    \begin{subfigure}[b]{0.49\textwidth}
        \centering
            \resizebox{\linewidth}{!}{ %
            \begin{tabular}{C{0.15\linewidth}C{0.15\linewidth}C{0.15\linewidth}C{0.15\linewidth}C{0.15\linewidth}C{0.15\linewidth}}
            \arrayrulecolor{vslamlab}
            \footnotesize ETH3D& \footnotesize HILTI  & \footnotesize EUROC & \footnotesize RGBDTUM  &  \footnotesize RGBDTUM \\
            \footnotesize L. Loop 1& \footnotesize Exp 04. & \footnotesize  MH04 Dif. & \footnotesize Fr.3 n.n.f &  \footnotesize Fr1. Desk2 \\
            \hline
            \\
            \multicolumn{5}{c}{\includegraphics[width=1\linewidth]{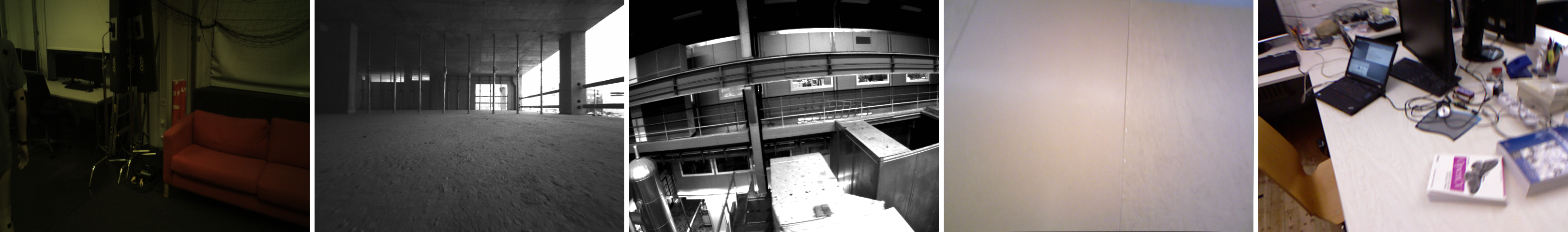}} \\
            \multicolumn{5}{c}{\includegraphics[width=1\linewidth]{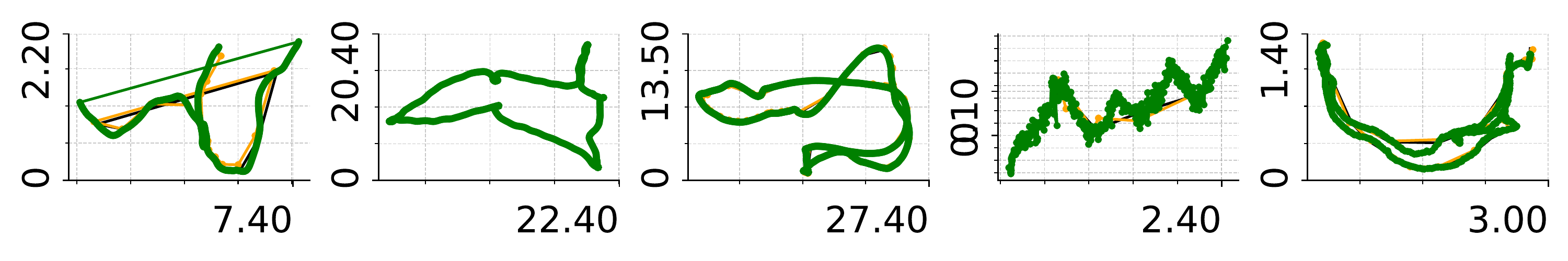}} \\
            \multicolumn{5}{c}{\includegraphics[width=1\linewidth]{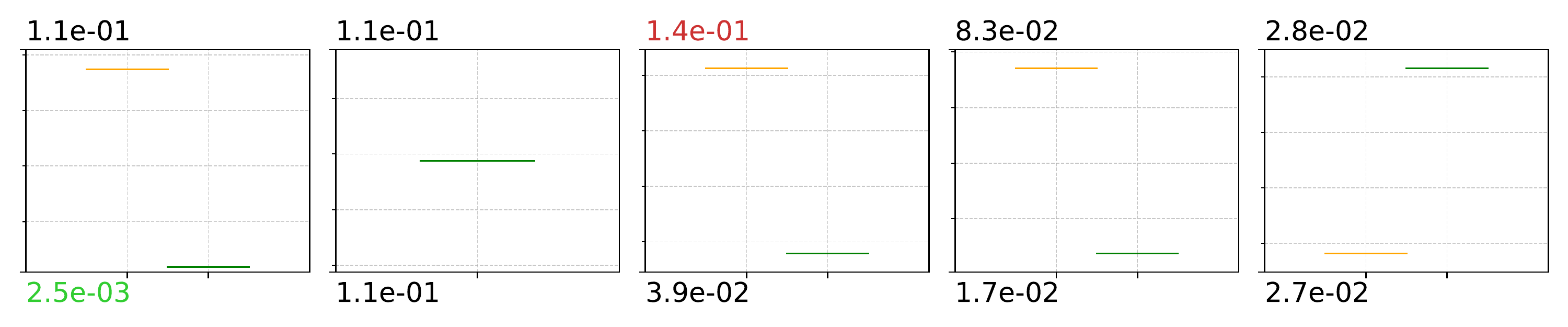}} \\
            \end{tabular}
            }
            \resizebox{0.65\linewidth}{!}{
            \begin{tabular}{ccccc}
            \myline{mast3r} MASt3R-SLAM & \myline{droid} DROID-SLAM
            \end{tabular}
            }
            \caption{\textbf{Difficult 2025}}
            \label{fig:configDifficult}
    \end{subfigure}
    \hfill
    \begin{subfigure}[b]{0.49\textwidth}
        \centering
        \resizebox{\linewidth}{!}{ %
        \begin{tabular}{C{0.15\linewidth}C{0.15\linewidth}C{0.15\linewidth}C{0.15\linewidth}C{0.15\linewidth}C{0.15\linewidth}}
        \arrayrulecolor{vslamlab}
        \footnotesize ETH3D& \footnotesize ETH3D & \footnotesize OPENLORIS & \footnotesize DRUNKARD  &  \footnotesize RGBDTUM \\
        \footnotesize Ceiling 2 & \footnotesize Einstein G. & \footnotesize Cafe 1  & \footnotesize 00000 &  \footnotesize Fr.3 Walk \\
        \hline
        \\
        \multicolumn{5}{c}{\includegraphics[width=1\linewidth]{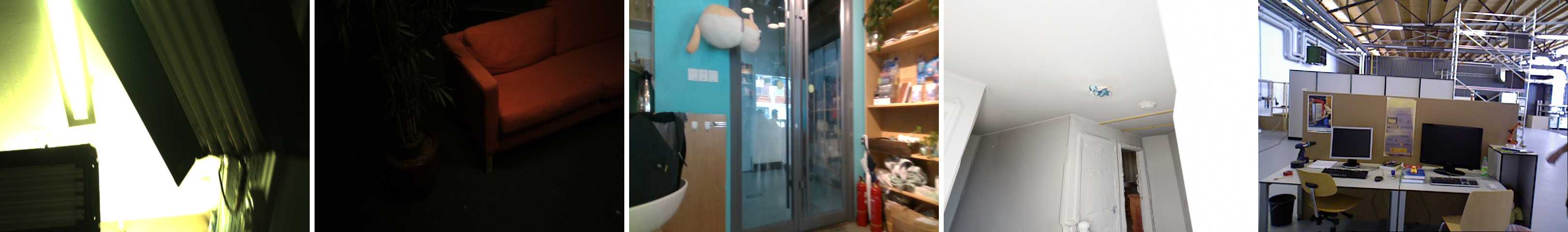}} \\
        \multicolumn{5}{c}{\includegraphics[width=1\linewidth]{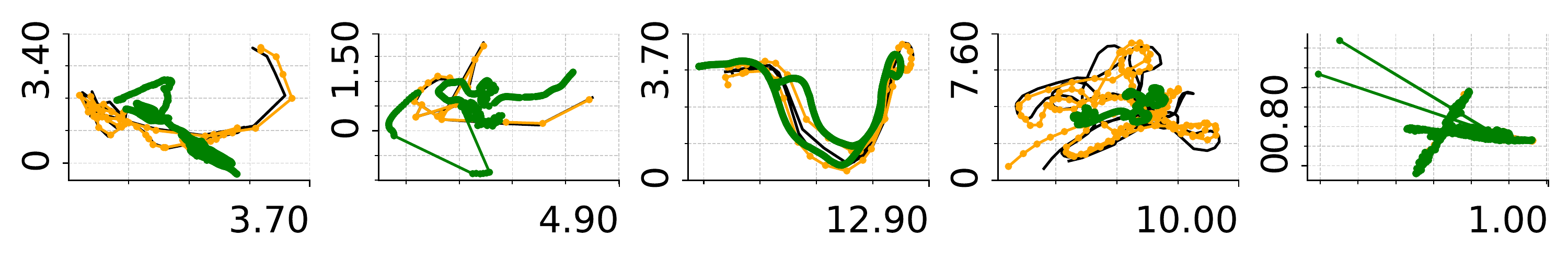}} \\
        \multicolumn{5}{c}{\includegraphics[width=1\linewidth]{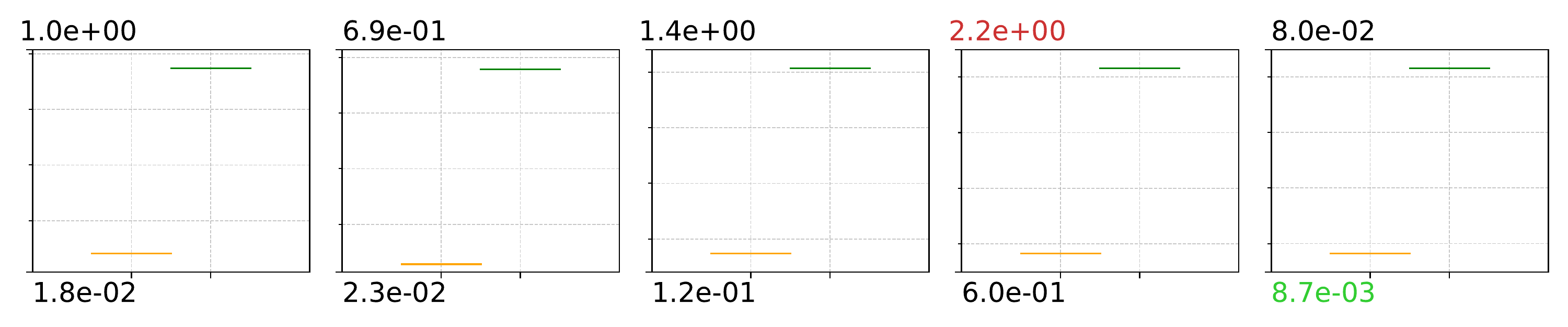}} \\
        \end{tabular}
        }
        \resizebox{0.65\linewidth}{!}{
        \begin{tabular}{ccccc}
        \myline{mast3r} MASt3R-SLAM & \myline{droid} DROID-SLAM
        \end{tabular}
        }
        \caption{\textbf{Extreme 2025}}
        \label{fig:configExtreme}
    \end{subfigure}
    
    \caption{\textbf{Benchmarking Configurations for VSLAM-LAB.} The figure presents four evaluation categories—Easy 2025, Medium 2025, Difficult 2025, and Extreme 2025—grouping sequences based on increasing levels of environmental complexity and motion challenges. Each configuration includes representative sequences (top row), camera trajectories (middle row), and ATE boxplots (bottom row) to facilitate performance comparisons across VSLAM methods. We note that these categories are examples only, and new categories can be easily added as described in Section~\ref{subset:experimentcustomization}.}
    \label{fig:main}
\end{figure*}

VSLAM-LAB automatically generates various plots and statistics to facilitate evaluation, with examples shown in Figures~\ref{fig:radar_all} and~\ref{fig:main}.

\textbf{ATE Boxplots:} Due to the non-deterministic nature of most VSLAM systems, multiple runs are necessary to account for random variations. To illustrate the variability in trajectory accuracy across all runs, ATE measurements are presented using boxplots. These plots provide a concise summary of absolute accuracy differences and system variability.

\textbf{Cumulative ATE Plot:} This plot represents the number of runs (y-axis) in which a system achieves an ATE below a given threshold (x-axis). The curve's position in the graph indicates the system's accuracy, while its width reflects variability. This visualization is particularly useful for comparing performance across different sequences~\cite{engel2016photometrically,engel2017direct}.

\textbf{Radar ATE Plots:} Given multiple ATE values from repeated runs of a method on a specific sequence, all ATE values are normalized using the median of the entire dataset. This normalization enables radar plots to represent the relative accuracy of different methods within a single visualization, even when applied to sequences with varying error scales. For instance, this approach allows meaningful comparisons between an indoor sequence with sub-centimeter accuracy and an outdoor sequence involving an autonomous vehicle. As shown in Figure~\ref{fig:radar_all}, these plots effectively highlight trends in method performance.

\textbf{Number of Estimated Frames:} The VSLAM-LAB evaluation also includes reporting the number of frames estimated by each method, the number of frames used to compute ATE, and the total number of frames in the sequence. Since not every frame has an associated ground truth due to factors such as motion capture system limitations or GPS failures, this metric ensures a fair comparison between methods. Some systems take a conservative approach by estimating only partial trajectories, while others estimate camera positions for a subset of keyframes or provide predictions for every frame.

\subsection{Other Tools} 

VSLAM-LAB provides evaluation tools to facilitate ablation studies, parameter search experiments, and analyses on specific or subsampled portions of sequences.

\textbf{Parameter Ablation:} The framework supports systematic experimentation with different parameter configurations, enabling seamless ablation studies:

\begin{lstlisting}
# VSLAMLAB/configs/exp_ablation.yaml

exp_ablation_mast3rslam:
  Config: config_easy.yaml
  NumRuns: 100
  Parameters: {verbose: 0, max_rgb: 120}
  Ablation: configs/ablation.csv
  Method: mast3rslam
\end{lstlisting}

A sample ablation file follows:

\begin{lstlisting}
# VSLAMLAB/configs/ablation.csv

exp_id  nFeatures  iniThFAST minThFAST
  0       500          20         7 
  1       500          20         10
  ...     ...          ...       ...
  99      1000         40         10
\end{lstlisting}

Similar to parameter ablation, noise ablation can be conducted by specifying a file that defines the type and magnitude of noise applied to experimental runs.

\textbf{RGB Sampling:} The experiment configuration file allows flexible frame selection policies. Users can specify particular sequence segments to process or downsample frames to a given frequency. This functionality helps to reduce computational load, decrease execution time, evaluate specific sequence portions, and assess system robustness under low-frame-rate conditions.

\section{EXPERIMENTS}
Different VSLAM methods exhibit varying performance depending on a wide range of factors, including scene characteristics (\eg, texture and depth distribution) and camera motion (\eg, translational and rotational velocities). Additionally, each method presents distinct trade-offs between accuracy, robustness, computational efficiency, and scalability.

Here, we introduce a set of four benchmarking configurations that cluster \numSequences{} sequences from \numDatasets{} diverse datasets based on the \textit{degree of difficulty} associated with the \textit{challenges} present in each scene. These new benchmarks enable straightforward comparison of existing techniques, and will help SLAM researchers to compare their new methods on established, standardized sequences. As methods become more robust and accurate, we can easily introduce new configurations with even more challenging and diverse datasets as described in Section~\ref{subsec:datasets}. Future works might also introduce domain-specific configurations, such as methods that focus on underwater SLAM or interplanetary SLAM. 

\textbf{Easy 2025:} The top row of Figure~\ref{fig:configEasy} displays frames from the five sequences included in the Easy 2025 configuration. These sequences feature well-structured environments with rich image textures and gentle motion. The main exception is sequence 04 from the KITTI dataset, which introduces large disparities due to the low frame rate (10 Hz) and vehicle movement. Aside from a distant moving car in the KITTI sequence, the scenes are predominantly static.

The ATE boxplot (third row) in Figure~\ref{fig:configEasy} highlights how, given the relatively low complexity of these sequences, traditional methods such as ORB-SLAM2 achieve competitive performance with state-of-the-art approaches like DROID-SLAM. Notably, ORB-SLAM2 outperforms both DROID-SLAM and MASt3R-SLAM in the KITTI dataset. This observation aligns with prior findings that these methods tend to degrade in performance in outdoor environments~\cite{teed2021droid}.

\textbf{Medium 2025:} Figure~\ref{fig:configMedium} includes sequences characterized by low-texture environments, such as the column in Fr3 Structure No Texture from the RGB-D TUM Benchmark and the Living Room Traj0 from the ICL-NUIM dataset. Other sequences, such as ME001 from the TartanAir dataset and a sequence from ScanNet++ feature stronger camera motion. DROID-SLAM successfully reconstructs the trajectories in these sequences, achieving

\textbf{Difficult 2025:} This configuration includes sequences characterized by strong camera motion, often leading to motion blur, such as the drone-mounted camera in sequence MH 04 Difficult from the EuRoC dataset and the handheld camera in sequence fr1 desk2 from the RGB-D TUM Benchmark. Additionally, it contains sequences with low structural complexity and poor texture, such as the construction site walls in sequence exp04 from the HILTI 2022 Challenge, as well as those captured under minimal lighting conditions, such as large loop 1 from the ETH3D Benchmark.

Figure~\ref{fig:configDifficult} presents only the performance of DROID-SLAM and MASt3R-SLAM, as DPVO and ORB-SLAM2 exhibited catastrophic accuracy degradation in these sequences. Notably, in challenging scenarios without extreme lighting variations, dynamic objects, or complete texture absence, DROID-SLAM consistently outperforms MASt3R-SLAM. It is important to note that MASt3R-SLAM is a relatively recent VSLAM system and has undergone less fine-tuning for these datasets.
 
\textbf{Extreme 2025:} Figure~\ref{fig:configExtreme} presents sequences that exhibit extreme conditions, including dynamic objects occupying a significant portion of the image, as seen in fr3 walking from the TUM-RGBD Benchmark. It also includes sequences with strong lighting variations, such as ceiling2 and einstein global light changes 1 from the ETH3D Benchmark, and scenes with an almost complete lack of texture, as observed in the walls of cafe1-1 from the OPENLORIS dataset or 00000 1014 level0 from the Drunkard's dataset.

In these very challenging sequences, MASt3R-SLAM demonstrates promising results, consistently achieving higher accuracy than DROID-SLAM.

\section{CONCLUSIONS}
As SLAM researchers, our focus should be on developing new methods that are robust to a wide range of scenarios and accurately track the camera's pose while building a map of the environment. While open-sourcing of SLAM methods (and computer vision / robotics papers in general) becomes more common, running these methods for benchmarking purposes and evaluating them on new datasets takes considerable time and effort.

In this paper, we introduce VSLAM-LAB, the first comprehensive framework for benchmarking VSLAM systems. VSLAM-LAB addresses the lack of standardization in VSLAM implementations, datasets, and evaluation procedures, which leads to fragmented comparisons and significant programming overhead. Our framework provides tools to unify VSLAM methods, data, and benchmarks, greatly simplifying development, reproducibility, and evaluation tasks. Installing and using VSLAM-LAB can be as easy as cloning our GitHub repository and running a single command: \emph{pixi run demo}.

Our experimental results compare four state-of-the-art VSLAM implementations across six sequences from four different datasets. For the first time, these comparisons highlight that ``old'' approaches like ORB-SLAM2 still outperforms new deep-learning techniques like DROID-SLAM and MASt3R-SLAM on some outdoor sequences, while the very recent MASt3R-SLAM performs best on extremely difficult sequences where ORB-SLAM2 and DPVO catastrophically fail. The inclusion of the recently proposed MASt3R-SLAM demonstrates VSLAM-LAB’s ability to seamlessly integrate novel implementations--an effort we will continue in the future, hopefully with the help of the research community.

\bibliographystyle{IEEEtran}
\bibliography{fontan}
\end{document}